\documentclass{article}

\usepackage{times}
\usepackage{graphicx} 
\usepackage{subfigure}

\usepackage{natbib}

\usepackage{algorithm}
\usepackage{algorithmic}

\usepackage{url}

\usepackage[accepted]{icml2019}

\usepackage{graphicx}

\usepackage{amsmath}
\usepackage{amsthm}
\usepackage{amssymb}
\usepackage{color}
\usepackage{subfigure}
\usepackage{multirow}
\usepackage{enumitem}

\icmltitlerunning{Efficient Nonconvex Regularized Tensor Completion
	with Structure-aware Proximal Iterations}

\usepackage{array}
\newcolumntype{L}[1]{>{\raggedright\let\newline\\\arraybackslash\hspace{0pt}}m{#1}}
\newcolumntype{C}[1]{>{\centering\let\newline  \\\arraybackslash\hspace{0pt}}m{#1}}
\newcolumntype{R}[1]{>{\raggedleft\let\newline \\\arraybackslash\hspace{0pt}}m{#1}}

\newtheorem{assumption}{Assumption}
\newtheorem{theorem}{Theorem}[section]
\newtheorem{lemma}[theorem]{Lemma}
\newtheorem{proposition}[theorem]{Proposition}
\newtheorem{corollary}[theorem]{Corollary}

\newtheorem{remark}{Remark}[section]
\newtheorem{definition}{Definition}

\newcommand{\NM}[2]{\left\|  #1 \right\|_{#2}}
\newcommand{\R}{\mathbb{R}}
\newcommand{\Px}[2]{\text{\normalfont prox}_{#1}(  #2 ) }

\newcommand{\Diag}[1]{ \text{\normalfont \,Diag} \left(  #1 \right) }

\newcommand{\Tr}[1]{\text{Tr}\left( #1 \right) }
\newcommand{\ip}[1]{{\left\langle  #1 \right\rangle}}

\usepackage[mathscr]{euscript}
\newcommand{\ten}[1]{ \boldsymbol{ \mathscr{ #1 } } }
\newcommand{\SO}[1]{P_{\Omega}\left( #1 \right) }

\newcommand{\mat}[1]{\text{\normalfont mat}( #1 )}

\newcommand{\vect}[1]{{\boldsymbol{\mathbf{#1}}}} 

\begin{document}

\twocolumn[
\icmltitle{Efficient Nonconvex Regularized Tensor Completion  \\
	with Structure-aware Proximal Iterations}

\icmlsetsymbol{equal}{*}

\begin{icmlauthorlist}
\icmlauthor{Quanming Yao}{4pa}
\icmlauthor{James T. Kwok}{ust}
\icmlauthor{Bo Han}{riken}
\end{icmlauthorlist}

\icmlaffiliation{4pa}{4Paradigm. Inc.}
\icmlaffiliation{ust}{Department of Computer Science and Engineering, 
	Hong Kong University of Science and Technology.}
\icmlaffiliation{riken}{RIKEN.}

\icmlcorrespondingauthor{Quanming Yao}{yaoquanming@4paradigm.com}

\icmlkeywords{}

\vskip 0.3in
]

\printAffiliationsAndNotice{}

\begin{abstract}
Nonconvex regularizers have been successfully used in
low-rank matrix
learning. 
In this paper,
we extend 
this to the more challenging problem of
low-rank tensor completion.
Based on the proximal average algorithm,
we develop an efficient solver
that avoids expensive 
tensor
folding and unfolding.
A special ``sparse plus
low-rank" structure, which is essential for fast computation of individual proximal
steps, 
is maintained
throughout the iterations.
We also incorporate adaptive momentum to further speed up empirical convergence.
Convergence results to critical points are provided
under smoothness and Kurdyka-Lojasiewicz conditions.
Experimental results on a number of synthetic and real-world data sets show that the proposed 
algorithm is more 
efficient in both
time and space,
and is also more accurate
than existing approaches.
\end{abstract}


\section{Introduction}
\label{sec:intro}

Tensors,
which can be seen as high-order matrices,
can be used to describe the multilinear relationships inside the data
\cite{Kolda2009,song2017tensor,papalexakis2017tensors}. They
are popularly used in areas such as computer vision \cite{vasilescu2002multilinear}, 
	recommender systems \cite{symeonidis2016matrix},  and
	signal processing \cite{cichocki2015tensor}.
Many of these are $3$-order tensors,
which are the focus in this paper.
Examples include
color images 
\cite{liu2013tensor}
and
hyperspectral images
\cite{signoretto2011tensor}.
In Youtube,
users can follow each other and can belong to the same subscribed channels.
By treating channels as the third dimension,
the users' co-subscription network can again be represented as a 3-order tensor \cite{lei2009analysis}.

In many applications, only a few tensor entries 
are observed.
For example,
each Youtube user often only interacts
with a few other users
\cite{lei2009analysis,davis2011multi}.
Tensor completion, which
aims at filling in this partially observed tensor,
has attracted a lot of interest
\cite{rendle2010pairwise,signoretto2011tensor,bahadori2014fast,cichocki2015tensor,symeonidis2016matrix,trouillon2017knowledge,lacroix2018canonical,wimalawarne2018efficient}.
In the related task of matrix completion, 
different rows/columns of the underlying full matrix often share similar characteristics, and
the matrix is thus 
low-rank
\cite{candes2009exact}.
The nuclear norm,
which is the tightest convex 
envelope of the rank \cite{boyd2009convex},
is popularly used 
as a surrogate for the matrix rank
in low-rank matrix completion \cite{cai2010singular,mazumder2010spectral}.


In tensor completion,
the low-rank assumption can also 
capture relatedness in the 
different tensor dimensions \cite{tomioka2010estimation,acar2011scalable,song2017tensor}.
However, tensors are more complicated
than matrices. Indeed,
even computation of the tensor rank is NP-hard \cite{hillar-13}.
In recent years, many convex relaxations
based on the matrix nuclear norm 
have been proposed
for tensors.
Examples include the tensor trace norm \cite{chandrasekaran2012convex},
overlapped nuclear norm \cite{tomioka2010estimation,gandy2011tensor},
and latent nuclear norm \cite{tomioka2010estimation}.
In particular,
the overlapped nuclear norm 
is the most popular, as it (i) 
can be computed exactly \cite{cheng2016scalable}, (ii)
has better low-rank approximation
\cite{tomioka2010estimation}, and (iii) can lead to
exact recovery \cite{tomioka2011statistical,tomioka2013convex,mu2014square}.

However,
the (overlapped) nuclear norm
equally penalizes all singular values.
Intuitively,
larger singular values are more informative and should be less
penalized \cite{mazumder2010spectral,lu2016nonconvex,yao2018large}.
In matrix completion,
various adaptive nonconvex regularizers have been recently introduced
to alleviate this problem.
Examples include
the capped-$\ell_1$ norm \cite{zhang2010analysis},
log-sum-penalty (LSP) \cite{candes2008enhancing},
truncated nuclear norm (TNN) \cite{hu2013fast},
smoothed-capped-absolute-deviation (SCAD) \cite{fan2001variable}
and
minimax concave penalty (MCP) \cite{zhang2010nearly}.
All these assign smaller penalties to the larger singular values.
This leads to better empirical performance \cite{lu2016nonconvex,gu2017weighted,yao2018large}
and statistical guarantee \cite{gui2016towards}.

Motivating by 
the success of adaptive nonconvex regularizers 
in matrix completion,
we propose to develop a nonconvex variant of the overlapped nuclear norm regularizer for tensor completion.
Unlike the standard convex tensor completion problem,
the resulting optimization problem is nonconvex and more difficult to solve.


Based on the proximal average algorithm 
\cite{bauschke2008proximal},
we develop 
in this paper
an efficient solver 
with much better time and space complexities.
The keys to  its success
are on (i) avoiding expensive tensor folding and unfolding,
(ii) maintaining a ``sparse plus low-rank'' structure on the iterates, and
(iii) incorporating 
the adaptive momentum \cite{Li2017ada}.
Convergence 
guarantees to critical points
are provided under smoothness and Kurdyka-Lojasiewicz \cite{attouch2013convergence} conditions.
Experiments 
on a number of synthetic and real-world data sets
show that the proposed algorithm is efficient
and has much better empirical performance than other low-rank tensor regularization and decomposition methods.



\textbf{Notation.}
Vectors
are denoted by lowercase boldface,
matrices
by uppercase boldface,
and
tensors
by boldface Euler.
For a matrix $\vect{A} \! \in \! \R^{m \! \times \! n}$
(assume $m \! \ge \! n$)
with singular values $\sigma_i$'s, its nuclear norm is $\| \vect{A} \|_* \! = \! \sum_i \! \sigma_i$.
For tensors,
we follow the
notation in \cite{Kolda2009}.
For a $3$-order tensor
$\ten{X} \! \in \! \R^{I_1\times I_2 \times I_3}$
(without loss of generality, we assume $\! I_1 \!\! \ge \!\! I_2 \!\! \ge \!\! I_3$),
its $(i_1, i_2, i_3)$th entry
is $\ten{X}_{i_1 i_2 i_3}$.
One can {\em unfold}
$\ten{X}$
along its $d$th-mode to
obtain the matrix $\vect{X}_{\left\langle  d \right\rangle} \!\! \in \!\! \R^{I_d \times
	(I_{\times} / I_d)}$,
whose $(i_d,j)$  entry
(where $\! j \!\! = \!\! 1 \! + \! \sum_{l=1,l\neq d}^3 (i_l \! - \! 1)\prod_{m=1, m\neq d}^{l-1}I_m$) is equal to
$\ten{X}_{i_1 i_2 i_3}$.
One can also {\em fold}  a matrix
$\vect{X}$
back to a tensor
$\ten{X} \! = \! \vect{X}^{\ip{d}}$,
such that $\ten{X}_{i_1 i_2 i_3} \! = \! \vect{X}_{i_d j}$,
and $j$ is as defined above.
The inner product of two tensors $\ten{X}$ and $\ten{Y}$ is $\ip{\ten{X} , \ten{Y}} \! = \!
\sum_{i_1=1}^{I_1} \sum_{i_2 = 1}^{I_2} \sum_{i_3 = 1}^{I_3} \ten{X}_{i_1 i_2 i_3} \ten{Y}_{i_1 i_2 i_3}$.
The Frobenius norm of $\ten{X}$ is $\|\ten{X}\|_F \! = \! \sqrt{\langle \ten{X}, \ten{X} \rangle}$.
For a proper and lower-semi-continuous function $f$,
$\partial f$ denotes its Frechet subdifferential
\cite{attouch2013convergence}.


\section{Related Works}
\label{sec:relworks}

\subsection{Low-Rank Matrix Learning} \vspace{-5px}
\label{sec:rel:nonreg}

Low-rank matrix learning can be formulated as:
\begin{align}
\min_{\mathbf{X}}
f(\mathbf{X}) + \lambda r( \mathbf{X} ),
\label{eq:lrmat}
\end{align}
where
$r$ is a low-rank regularizer,
$\lambda \ge 0$ is a hyper-parameter,
$f$ is a 
$\rho$-Lipschitz smooth\footnote{In other words, $\| \nabla f(\mathbf{X}) \! - \! \nabla f(\vect{Y}) \|_F^2 
	\! \le \! \rho \NM{\mathbf{X} \! - \! \vect{Y}}{F}^2$ for any $\mathbf{X}$ and $\vect{Y}$.}
loss.
A common choice of $r$ 
is the nuclear norm.
Using the 
proximal algorithm \cite{parikh2013proximal},
the iterate is given by
$\mathbf{X}_{t+1} = \Px{\frac{\lambda}{\tau} \|\cdot\|_*}{ \mathbf{Z}_t}$,
where 
\begin{equation} \label{eq:zt21}
\mathbf{Z}_t = \mathbf{X}_t - \frac{1}{\tau} \nabla f(\mathbf{X}_t),
\end{equation} 
$\tau > \rho$ is the stepsize, and
\begin{align}
\Px{\frac{\lambda}{\tau} \NM{\cdot}{*}}{\mathbf{Z}}
\equiv \text{\normalfont Arg}\min_{\mathbf{X}} \frac{1}{2} \NM{\mathbf{X} - \mathbf{Z}}{F}^2  
+ \frac{\lambda}{\tau} \NM{\mathbf{X}}{*},
\label{eq:proxsvt}
\end{align}
is the proximal step.
The following Lemma shows that
it can be obtained from the SVD of $\mathbf{Z}$.
Note that 
shrinking of
the singular values 
encourages
$\mathbf{X}_t$ to be low-rank.

\begin{lemma}{\normalfont \cite{cai2010singular}}
$\Px{\lambda\NM{\cdot}{*}}{\mathbf{Z}} = \mathbf{U}
(\mathbf{\Sigma} - \lambda \mathbf{I})_+ \mathbf{V}^{\top}$,
where 
$\mathbf{U} \mathbf{\Sigma} \mathbf{V}^{\top}$ is the SVD of $\mathbf{Z}$,
and
$\left[ (\mathbf{X})_+ \right]_{ij} = \max(\mathbf{X}_{ij}, 0)$.
\label{lem:svt}
\end{lemma}


\subsubsection{Matrix Completion}

A special class of low-rank matrix learning problems is
matrix completion, which attempts to find a low-rank matrix that agrees with the observations
in data $\mathbf{O}$:
\begin{align}
\min_{\mathbf{X} \in \R^{m \times n}}
\frac{1}{2}\NM{P_{\Omega}(\mathbf{X} - \mathbf{O})}{F}^2
+ \lambda \NM{\mathbf{X}}{*}.
\label{eq:matcomp}
\end{align}
The positions of observed elements 
in $\mathbf{O}$
are indicated by $1$'s
in the binary matrix $\Omega$,
$[ \SO{\mathbf{X}} ]_{i j} = \mathbf{X}_{i j}$ if $\Omega_{i j} = 1$ and $0$ otherwise.
Setting
$f(\mathbf{X})=
\frac{1}{2}\NM{P_{\Omega}(\mathbf{X} - \mathbf{O})}{F}^2$
in 
(\ref{eq:lrmat}),
$\vect{Z}_t$ in (\ref{eq:zt21}) becomes:
\begin{align}
\label{eq:zt1}
\vect{Z}_t
=
\vect{X}_{t} - \frac{1}{\tau} P_{\Omega}( \vect{X}_{t} - \vect{O} ).
\end{align}
Note that
$\mathbf{Z}_t$ has a ``sparse plus low-rank'' structure,
with 
a low-rank 
$\mathbf{X}_t$ 
and 
sparse
$\frac{1}{\tau} P_{\Omega}( \vect{X}_{t} - \vect{O} )$.
This allows
the SVD  computation
in Lemma~\ref{lem:svt} 
to be 
much more 
efficient \cite{mazumder2010spectral}.
Specifically, 
on using the power method
to compute the SVD of $\vect{Z}_t$, most effort is spent on 
multiplications of the form $\vect{Z}_t \mathbf{b}$ (where $\mathbf{b} \in
\R^{m}$)
and $\mathbf{a}^{\top} \vect{Z}_t$ (where $\mathbf{a} \in \R^{n}$).
Let 
$\mathbf{X}_{t}$ 
in (\ref{eq:zt1})
be low-rank factorized as
$\mathbf{U}_{t} \mathbf{V}_{t}^{\top}$ with rank
$k_{t}$.
Computing
\begin{align}
\vect{Z}_t \mathbf{b}
= 
\mathbf{U}_{t} ( \mathbf{V}_{t}^{\top} \mathbf{b}  )
- \frac{1}{\tau} P_{\Omega} ( \vect{Y}_{t} - \mathbf{O} ) \mathbf{b}
\label{eq:zt1v}
\end{align}
takes $O( (m + n) k_t + \NM{\Omega}{1})$ time.
Usually, $k_{t} \ll n$ and $\NM{\Omega}{1} \ll m n$.
This is much faster than direct multiplying $\vect{Z}_t$ and $\mathbf{b}$,
which takes $O( m n )$ time.
The same holds for $\mathbf{a}^{\top} \vect{Z}_t$.
Thus,
the proximal step in \eqref{eq:proxsvt} 
take $O( (m + n) k_t k_{t + 1} + \NM{\Omega}{1} k_{t + 1} )$ time,
while a direct computation
without utilizing the ``sparse plus low-rank'' structure
takes $O( m n k_{t + 1} )$ time.
Besides, as
only 
$\SO{\mathbf{X}_{t}}$ 
and 
the factorized form of $\mathbf{X}_{t}$ 
need to be kept,
the space complexity is reduced from $O( m n )$ to
$O( (m + n) k_t + \NM{\Omega}{1} )$.


\subsubsection{Nonconvex Low-Rank Regularizer}

Instead of using a convex $r$ in \eqref{eq:lrmat}, the following
nonconvex regularizer 
is commonly used \cite{gui2016towards,lu2016nonconvex,gu2017weighted,yao2018large}:
\begin{align}
\phi(  \mathbf{X} )
= \sum_{i = 1}^{n} \kappa( \sigma_i( \mathbf{X} ) ),
\label{eq:lrphi}
\end{align}
where $\kappa$ is nonconvex and possibly nonsmooth.
We assume the following on $\kappa$.



\begin{assumption}\label{ass:kappa}
	$\kappa(\alpha)$ is a concave,
	nondecreasing
	and $L$-Lipschitz continuous
	function on $\alpha \ge 0$
	with $\kappa(0) = 0$.
\end{assumption}
Examples
of $\kappa$ include the
capped-$\ell_1$ penalty:
$\kappa(\sigma) = \log (\frac{1}{\theta} \sigma + 1)$ 
\cite{zhang2010analysis},  
and log-sum-penalty (LSP):
$\kappa(\sigma) = \min(\sigma, \theta )$
(where $\theta > 0 $ is a constant)
\cite{candes2008enhancing}.
More can be found in Appendix~\ref{app:egreg}.
They have similar statistical guarantees \cite{gui2016towards}, and 
empirically perform better than the convex nuclear norm
\cite{lu2016nonconvex,yao2018large}.

The proximal algorithm can 
again
be used,
and converges to a critical point \cite{attouch2013convergence}.
Analogous to Lemma~\ref{lem:svt},
the underlying proximal step 
\begin{align}
\Px{\frac{\lambda}{\tau} \phi}{\mathbf{Z}}
\equiv \text{\normalfont Arg}\min_{\mathbf{X}} \frac{1}{2} \NM{\mathbf{X} - \mathbf{Z}}{F}^2  
+ \frac{\lambda}{\tau} \phi(\mathbf{X}),
\label{eq:proxgsvt}
\end{align}
can be obtained as follows.

\begin{lemma}{\normalfont \cite{lu2016nonconvex}}
\label{lem:gsvt}
$\Px{\lambda\phi}{\mathbf{Z}} = \mathbf{U} \Diag{  y_1, \dots, y_{n} }
\mathbf{V}^{\top}$, where
$\mathbf{U} \mathbf{\Sigma} \mathbf{V}^{\top}$ is the SVD of $\mathbf{Z}$, and
$y_i = \arg \min_{y \ge 0} \frac{1}{2} (y - \sigma_i(\mathbf{Z}))^2 + \lambda \kappa(y)$.
\end{lemma}


\subsection{Low-Rank Tensor Learning} \vspace{-5px}
\label{sec:ten:reg}

Recently, the nuclear norm regularizer has been extended to tensors.
Here, we focus on 
the 
overlapped nuclear norm
\cite{tomioka2010estimation,gandy2011tensor}, and its nonconvex extension that will
be introduced in Section~\ref{sec:proalg}.


\begin{definition}
\label{def:tensor_norm}
For a
$M$-order tensor $\ten{X}$, 
the {\em overlapped nuclear norm \/} is $\| \ten{X} \|_\text{overlap} \! = \! \sum_{m=1}^M  \lambda_m \|
\ten{X}_{\ip{m}} \|_*$, where $\{\lambda_m \! \ge \! 0\}$ are hyperparameters.
\end{definition}

Factorization methods,
such as the Tucker/CP \cite{Kolda2009} and tensor-train decompositions \cite{oseledets2011tensor},
have also been used for
low-rank tensor
learning. 
Compared to nuclear norm regularization,
they usually offer worse approximations 
and inferior performance
\cite{tomioka2011statistical,liu2013tensor,guo2017efficient}.


\section{Nonconvex Low-Rank Tensor Completion}
\label{sec:proalg}

By integrating a nonconvex $\phi$ 
in \eqref{eq:lrmat}
with the overlapped nuclear norm, 
the tensor completion problem 
becomes
\begin{align}
\min_{\ten{X}}
F(\ten{X}) \equiv
\frac{1}{2}\NM{P_{\Omega}(\ten{X} - \ten{O})}{F}^2
+  \sum_{d = 1}^D \frac{\lambda_d}{D} \phi( \ten{X}_{\ip{d}} ).
\label{eq:pro}
\end{align}
Note that  
we only sum over
$D \! \le \! M$ modes.
This is useful 
as 
the third mode 
is 
sometimes 
already small
(e.g., the number of bands in images), and
so does not need to be low-rank regularized.
When $D \! = \! 1$, \eqref{eq:pro} reduces to 
the matrix completion problem 
$\min_{\mathbf{X} \in \R^{I_1 \times I_2 I_3}}  
\frac{1}{2} \| P_{\Omega}(\mathbf{X} - \ten{O}_{\ip{1}}) \|_F^2 + \lambda_1
\phi( \mathbf{X} )$,
which can be solved by the proximal algorithm as in \cite{lu2016nonconvex,yao2018large}.  
In the sequel, we only consider $D \! \neq \! 1$.

When $\kappa(\alpha) \! = \! |\alpha|$,
\eqref{eq:pro} reduces to  
(convex) overlapped nuclear norm regularization. 
While $D$ may not be equal to  $M$, it can be easily shown that
optimization solvers such as alternating direction of multiple multipliers (ADMM)
and fast low-rank tensor completion (FaLRTC)
can still be used as in
\cite{tomioka2010estimation,liu2013tensor}.
However,
when $\kappa$
is nonconvex,
ADMM no longer guarantees convergence, and
FaLRTC's
dual 
cannot be 
derived.



\subsection{Structure-aware Proximal Iterations} \vspace{-5px}
\label{ssec:proxavg}

As in Section~\ref{sec:rel:nonreg},
we 
solve \eqref{eq:pro}
with the proximal algorithm.
However,
the proximal step for $\sum_{d = 1}^D \lambda_i \phi( \ten{X}_{\ip{d}} )$ is 
not simple.
To address this problem, we use
the proximal average (PA) algorithm 
\cite{bauschke2008proximal,yu2013better}.

Let $\mathcal{H}$ be a Hilbert space.
Consider the following problem with $K$
possibly nonsmooth regularizers,
whose individual proximal steps are assumed to be easily computable.
\begin{align}
\min_{\ten{X} \in \mathcal{H}} f(\ten{X}) +  \sum_{i = 1}^K \frac{\lambda_i}{K} g_i(\ten{X}),
\label{eq:compopt}
\end{align}
where
$f$ is 
convex
and 
Lipschitz-smooth,
while each $g_i$ is convex but possibly nonsmooth.
The PA algorithm generates the iterates $\ten{X}_{t}$'s as
\vspace{-4px}
\begin{align}
\ten{X}_{t}
& = \frac{1}{K} \sum_{i = 1}^K
 \ten{Y}_{t}^i,
\label{eq:proxavg1}
\\
\ten{Z}_t 
& =\ten{X}_t - \frac{1}{\tau} \nabla f(\ten{X}_t), 
\label{eq:proxavg2} 
\\
\ten{Y}_{t + 1}^i
& = \Px{\frac{\lambda_i}{\tau} g_i}{\ten{Z}_t},
\quad i = 1, \dots, K.
\label{eq:proxavg3} 
\end{align}
Recently,
the PA algorithm is also extended to nonconvex $f$ and $g_i$'s,
where each $g_i$
admits a difference-of-convex decomposition\footnote{In other words, each $g_i$ can be decomposed as $g_i = \bar{g}_i - \hat{g}_i$
	where $\bar{g}_i$ and $\hat{g}_i$ are two convex functions.}
\cite{yu2015minimizing,zhong2014gradient}.

Note that 
\eqref{eq:pro} is of the form in 
\eqref{eq:compopt}, and 
$\phi$ in \eqref{eq:lrphi} 
admits a difference-of-convex decomposition
\cite{yao2018large},
the PA algorithm can be used to
generate
the iterates 
as:
\vspace{-6px}
\begin{eqnarray}
\ten{X}_{t} & = & \frac{1}{D} \sum_{i = 1}^D \ten{Y}^i_{t},
\label{eq:tsplr_1}
\\
\ten{Z}_t
& = &
\ten{X}_t - \frac{1}{\tau} \SO{\ten{X}_t - \ten{O}}.
\label{eq:tsplr_2}
\end{eqnarray}
However,
as the regularizer $\phi$ is imposed on the unfolded matrix $\ten{X}_{\ip{d}}$, not on $\ten{X}$ directly,
the proximal steps need to be performed as follows.

\begin{proposition}\label{pr:pxavg}
For problem \eqref{eq:pro},
step~\eqref{eq:proxavg3} of the PA algorithm
can be performed as
\begin{eqnarray}
\ten{Y}^i_{t + 1}
& = & 
\left[ \Px{\frac{\lambda_i}{\tau} \phi}{ [\ten{Z}_t]_{\ip{i}} } \right]^{\ip{i}}.
\label{eq:tsplr_3}
\end{eqnarray}
\end{proposition}
The individual proximal steps in \eqref{eq:tsplr_3} can be computed using Lemma~\ref{lem:gsvt} based on SVD.
However, tensor folding and unfolding are required in \eqref{eq:tsplr_3}.
A direct implementation  
takes $O(I_{\times})$ space and $O( I_{\times} I_{+}  )$ time per iteration, where
$I_{\times} = \prod_{i = d}^3 I_d$ and $I_{+} = \sum_{i = d}^3 I_d$,
and is expensive.
In the following, 
we show how the PA iterations
can be computed efficiently by utilizing the 
``sparse plus low-rank'' structures.


\subsubsection{Keeping the Low-Rank Factorizations} 

In (\ref{eq:tsplr_3}),
let 
$\vect{Y}^i_{t + 1} = \Px{\frac{\lambda_i}{\tau} \phi}{ \vect{Z}^i_t }$, where
$\vect{Z}^i_t = \left[ \ten{Z}_t \right]_{\ip{i}}$.
Recall that $\vect{Y}_t^i$ 
is low-rank.
Let
its rank be
$k^i_t$. 
In each iteration,
we avoid 
constructing the dense 
$\ten{Y}_t^i$ 
by storing 
$\vect{Y}_t^i$ as
$\mathbf{U}_{t}^i  (\mathbf{V}_{t}^i)^{\top}$, 
where $\mathbf{U}_{t}^i \in \R^{I_i \times k_t^i}$
and $\mathbf{V}_{t}^i \in \R^{(I_{\times}\!/I_i) \times k_t^i}$.
We also avoid getting 
$\ten{X}_t$ 
in \eqref{eq:tsplr_1}
by storing 
it implicitly 
as 
\vspace{-6px}
\begin{align}
\ten{X}_t = \frac{1}{D}
\sum_{i = 1}^D \left( \mathbf{U}_{t}^i ( \mathbf{V}_{t}^i )^{\top} \right)^{\ip{i}}.
\label{eq:lowrf}
\end{align}

\subsubsection{Maintaining ``Sparse plus Low-Rank''}

Using (\ref{eq:lowrf}),
$\ten{Z}_t$ in \eqref{eq:tsplr_2} can be rewritten as
\vspace{-6px}
\begin{align}
\ten{Z}_t
=
\frac{1}{D}
\sum_{i = 1}^D ( \mathbf{U}_{t}^i  (\mathbf{V}_{t}^i)^{\top} )^{\ip{i}}
- \frac{1}{\tau} \SO{\ten{X}_t \! - \! \ten{O}}.
\label{eq:zt}
\end{align}
The sparse tensor $\SO{\ten{X}_t \! - \! \ten{O}}$ 
can be 
constructed
efficiently 
by using the coordinate format\footnote{For a sparse 3-order tensor, 
its $p$th nonzero element is represented 
in the coordinate format
as $(i_p^1, i_p^2, i_p^3, v_p)$,
	where $i_p^1, i_p^2, i_p^3$ are indices on each mode and $v_p$ is the value.  Using \eqref{eq:lowrf},
	$v_p$ 
of $\SO{\ten{X}_t \! - \! \ten{O}}$ 
	can be computed 
	by finding the corresponding rows in 
	$\mathbf{U}_{t}^i$ and $\mathbf{V}_{t}^i$,
	which takes $O( \sum_{i = 1}^D k_t^i )$ time.}
\cite{bader2007efficient}.
As $\sum_{i = 1}^D ( \mathbf{U}_{t}^i  (\mathbf{V}_{t}^i)^{\top} \! )^{\ip{i}}$
is a sum of tensor
(folded from low-rank matrices)
and $\frac{1}{\tau} \SO{\ten{X}_t \! - \! \ten{O}}$ is sparse,
$\ten{Z}_t$ is
also ``sparse plus low-rank''.


Recall that
the proximal step   in (\ref{eq:tsplr_3})
requires SVD,
which involves matrix multiplications of the form
$( \ten{Z}_t )_{\ip{i}} \mathbf{b}$ (where
$\mathbf{b} \! \in \! \R^{I_{\times}\!/I_i}$)
and
$\mathbf{a}^{\top} ( \ten{Z}_t )_{\ip{i}}$ (where $\mathbf{a} \! \in \! \R^{I_i}$).
Using
the ``sparse plus low-rank'' structure in
(\ref{eq:zt}),
\begin{eqnarray}
( \ten{Z}_t )_{\ip{i}} \mathbf{b} \!\!\!\!
& \! = \! & \!\!\!\! \frac{1}{D} \mathbf{U}_{t}^i [ (\mathbf{V}_{t}^i)^{\top} \mathbf{b} ]
\! + \!  
\frac{1}{D} \! \sum_{j \neq i}
[ ( \mathbf{U}_{t}^j (\mathbf{V}_{t}^j)^{\top} )^{\ip{j}} ] _{\ip{i}}
\mathbf{b}
\notag
\\
&
& \!\!\!\! - \frac{1}{\tau} [ \SO{\ten{X}_t \! - \! \ten{O}} ]_{\ip{i}} \mathbf{b},
\label{eq:tenztv}
\end{eqnarray}
\begin{eqnarray}
\!\!
\mathbf{a}^{\top} \! ( \ten{Z}_t )_{\ip{i}} \!\!\!\!
& \! = \! & \!\!\!\!
\frac{1}{D} (\mathbf{a}^{\top} \! \mathbf{U}_{t}^i) (\mathbf{V}_{t}^i)^{\top}
\!\!\! + \!
\frac{1}{D} \! \sum_{j \neq i} \!
\mathbf{a}^{\top}
\!
[( \mathbf{U}_{t}^j (\mathbf{V}_{t}^j)^{\top} )^{\ip{j}} ]_{\ip{i}}
\notag
\\
&& \!\!\!\!\!
- \frac{1}{\tau} \mathbf{a}^{\top} \!\! \left[ \SO{\ten{X}_t \! - \! \ten{O}} \right]_{\ip{i}}.
\label{eq:tenztu}
\end{eqnarray}
The first term in both (\ref{eq:tenztv}) and (\ref{eq:tenztu}) can be easily computed
in $O((I_{\times} /I_i \! + \! I_i) k_t^i)$ space and time.
$[ \SO{\ten{X}_t \! - \! \ten{O}} ]_{\ip{i}} \mathbf{b}$ and
$\mathbf{a}^{\top} \left[ \SO{\ten{X}_t \! - \! \ten{O}} \right]_{\ip{i}}$
are sparse.
Using sparse tensor packages
such as the Tensor Toolbox \cite{bader2007efficient},
$[ \SO{\ten{X}_t - \ten{O}} ]_{\ip{i}} \mathbf{b}$ and $\mathbf{a}^{\top} [ \SO{\ten{X}_t - \ten{O}} ]_{\ip{i}}$
can be computed
in $O( \NM{\Omega}{1} )$ space and time.

Computing
$\mathbf{a}^{\top} [ ( \mathbf{U}_t^i (\mathbf{V}_t^i)^{\top} )^{\ip{j}} ]_{\ip{i}}$
and $[  ( \mathbf{U}_t^i (\mathbf{V}_t^i)^{\top} )^{\ip{j}}  ]_{\ip{i}} \mathbf{b}$
in (\ref{eq:tenztv}), (\ref{eq:tenztu}) 
involves folding/unfolding and is expensive.
By examining how elements are ordered by folding and unfolding,
the following shows that
$\mathbf{a}^{\top} [ ( \mathbf{U}_t^i (\mathbf{V}_t^i)^{\top} )^{\ip{j}} ]_{\ip{i}}$
and $[  ( \mathbf{U}_t^i (\mathbf{V}_t^i)^{\top} )^{\ip{j}}  ]_{\ip{i}} \mathbf{b}$
can be reformulated without 
explicit folding / unfolding, and thus be
computed more efficiently.

\begin{proposition} \label{pr:mulv}
Let $\mathbf{U} \in \R^{I_{i} \times k}$, $\mathbf{V} \in \R^{I_{\times}\!/I_i \times k}$,
and $\mathbf{u}_p$ (resp.$\mathbf{v}_p$) be the $p$th column of $\mathbf{U}$ (resp.$\mathbf{V}$).
For any $\mathbf{a} \in \R^{I_{j}}$
and $\mathbf{b} \in \R^{I_{\times}\!/I_{j}}$,
we have
\begin{align*}
\mathbf{a}^{\top}
[  ( \mathbf{U} \mathbf{V}^{\top} )^{\ip{i}} ]_{\ip{j}}
& =
\sum_{p = 1}^k
\mathbf{u}_p^{\top} \otimes
[  \mathbf{a}^{\top} \mat{\mathbf{v}_p} ],
\\
[ (\mathbf{U} \mathbf{V}^{\top} )^{\ip{i}} ]_{\ip{j}}
\mathbf{b}
& =
\sum_{p = 1}^k
\mat{ \mathbf{v}_p } \mat{ \mathbf{b} }^{\top} \mathbf{u}_p,
\end{align*}
where $\!\otimes\!$ is the Kronecker product,
and $\mat{\mathbf{c}}$ reshapes
vector 
$\mathbf{c} \! \in \! \R^{I_{i} I_{j}}$ into a $I_{i} \! \times \! I_{j}$ matrix.
\end{proposition}

\begin{table*}[ht]
	\centering
	\vspace{-10pt}
	\renewcommand{\arraystretch}{1.3}
	\caption{Comparison of the proposed NORT (Algorithm~\ref{alg:tcnr}) and direct implementations of the PA algorithm.}
	\scalebox{0.90}{
		\begin{tabular}{c | C{220px} | c | c}
			\hline
			& per-iteration time complexity         & space           & convergence \\ \hline
			direct & $O( I_{\times} \sum_{i = 1}^D I_i  )$ & $O(I_{\times})$ & slow        \\ \hline
			NORT  & $O( \sum_{i = 1}^D \sum_{j \neq i} (\frac{1}{I_i} \! + \! \frac{1}{I_j}) k_t^i k_{t +
				1}^i I_{\times} \! + \! \NM{\Omega}{1} (k_{t}^i + k_{t + 1}^i) )$                                      & $O( \sum_{i = 1}^D \sum_{j \neq i} (\frac{1}{I_i} \! + \! \frac{1}{I_j}) k_t^i
			I_{\times} \! + \! \NM{\Omega}{1} )$                &  fast           \\ \hline
	\end{tabular}}
	\label{tab:overview}
	\vspace{-10pt}
\end{table*}

\begin{remark}
As a special case,
take $i \! = \! 1$ and $I_j = 1$ where $j \! \in \! \{ 2, 3 \}$,
the $I_1 \! \times\! I_2 \! \times\! I_3$ tensor reduces to a matrix.
Proposition~\ref{pr:mulv} then becomes
$\mathbf{a}^{\top}
[ ( \mathbf{U} \mathbf{V}^{\top} )^{\ip{1}} ]_{\ip{j}}
\! = \! \sum_{p = 1}^k
\mathbf{u}_p^{\top}(\mathbf{a}^{\top}\mathbf{v}_p)
\! = \! (\mathbf{a}^{\top} \mathbf{U}) \mathbf{V}^{\top},$
$[  (\mathbf{U} \mathbf{V}^{\top} )^{\ip{1}} ]_{\ip{j}}
\mathbf{b}
\! = \! \sum_{p = 1}^k
\mathbf{v}_p (\mathbf{b}^{\top} \mathbf{u}_p)
\! = \! \mathbf{U} (\mathbf{V}^{\top} \mathbf{b})$,
and
\eqref{eq:tenztv} 
reduces to \eqref{eq:zt1v}.
\end{remark}

Computation of $\mathbf{a}^{\top} [ ( \mathbf{U}_t^i (\mathbf{V}_t^i)^{\top} )^{\ip{j}} ]_{\ip{i}}$
takes
a total of
$O((\frac{1}{I_i} \! + \! \frac{1}{I_j}) k_t^i I_{\times} )$ time 
and $O( (\frac{1}{I_i} \! + \! \frac{1}{I_j}) I_{\times})$ space.
The same holds for computation of $[  ( \mathbf{U}_t^i (\mathbf{V}_t^i)^{\top} )^{\ip{j}}  ]_{\ip{i}} \mathbf{b}$.
This is much less expensive than direct evaluation, which takes
$O( k_t^i I_{\times} )$ time
and
$O( I_{\times} )$ space.

Combining the above,
and noting that we have to keep the factorized form $\mathbf{U}_t^i
(\mathbf{V}_t^i)^{\top}$ of $\vect{Y}_t^i$,
computing proximal steps in \eqref{eq:tsplr_3} takes
$O( \sum_{i = 1}^D \sum_{j \neq i} (\frac{1}{I_i} \! + \! \frac{1}{I_j}) k_t^i
I_{\times} \! + \! \NM{\Omega}{1} )$
space
and
$O( \sum_{i = 1}^D \sum_{j \neq i} (\frac{1}{I_i} \! + \! \frac{1}{I_j}) k_t^i k_{t +
	1}^i I_{\times} \! + \! \NM{\Omega}{1} (k_{t}^i \! + \! k_{t + 1}^i) )$
time.

\begin{remark}
In tensor applications such as 
tensor regression \cite{signoretto2014learning} and cokriging \cite{bahadori2014fast},
terms of the form
$\mathbf{a}^{\top}
[  ( \mathbf{U} \mathbf{V}^{\top} )^{\ip{i}} ]_{\ip{j}}$ and
$[ (\mathbf{U} \mathbf{V}^{\top} )^{\ip{i}} ]_{\ip{j}} \mathbf{b}$ are also involved.
Thus, Proposition~\ref{pr:mulv} can also be used for speedup.
Details are in Appendix~\ref{app:otherprapp}.
\end{remark}

\subsubsection{Complexities}

In each PA iteration,
$D$ proximal steps
are performed
in \eqref{eq:tsplr_3}.
The whole PA algorithm thus takes a total of
$O( \sum_{i = 1}^D \sum_{j \neq i} (\frac{1}{I_i} \! + \! \frac{1}{I_j}) k_t^i
I_{\times} + \NM{\Omega}{1} )$
space
and
$O( \sum_{i = 1}^D \sum_{j \neq i} (\frac{1}{I_i} \! + \! \frac{1}{I_j}) k_t^i k_{t +
	1}^i I_{\times} + \NM{\Omega}{1} (k_{t}^i + k_{t + 1}^i) )$
time for each iteration.
As $k_t^i$, $k_{t + 1}^i \ll I_i$, 
these are much lower than those of a direct implementation
(Table~\ref{tab:overview}).
Moreover, the PA algorithm has a convergence rate of $O(1/T)$, where $T$ is the number of iterations
\cite{zhong2014gradient,yu2015minimizing}.



\subsection{Use of Adaptive Momentum}
\label{ssec:fastconv}

The PA algorithm uses only first-order information,
and empirically can be slow to converge \cite{parikh2013proximal}.
To address
this problem,
we adopt adaptive momentum,
which has been popularly used for stochastic gradient descent
\cite{duchi2011adaptive,kingma2014adam}
and proximal algorithms
\cite{li2015accelerated,yao2017efficient,Li2017ada}.
The idea is to use historical iterates to speed up convergence.
Here, 
we adopt the adaptive scheme in \cite{Li2017ada}.
The resultant procedure, 
shown in Algorithm~\ref{alg:tcnr},
will be called
\underline{NO}ncvx \underline{R}egularized \underline{T}ensor  (NORT).
Note that 
even when step~6 is performed,
$\ten{Z}_t$ in step~10
still has the ``sparse plus low-rank'' structure on $\ten{Z}_t$,
since
$\ten{Z}_t
=
\frac{1 + \gamma_t}{D} \sum_{i = 1}^D ( \mathbf{U}_{t}^i  (\mathbf{V}_{t}^i)^{\top} )^{\ip{i}}
-  \frac{\gamma_t}{D} \sum_{i = 1}^D ( \mathbf{U}_{t - 1}^i  (\mathbf{V}_{t - 1}^i)^{\top} )^{\ip{i}}
- \frac{1}{\tau} \SO{\bar{\ten{X}}_t - \ten{O}}$.
The resultant time and space complexities are the same
as in Section~3.1.3.


\begin{algorithm}[ht]
\caption{\underline{NO}nconvex \underline{R}egularized \underline{T}ensor (NORT).}
\begin{algorithmic}[1]
	\STATE initialize $\ten{X}_0 \! = \! \ten{X}_1 \! = \! 0$,
	$\tau > \rho + D L$
	and $\gamma_1, p \in (0, 1)$;
	\FOR{$t = 1, \dots, T$}
	\STATE $\ten{X}_{t + 1} = \frac{1}{D} \sum_{i = 1}^D ( \mathbf{U}_{t + 1}^i ( \mathbf{V}_{t + 1}^i )^{\top} )^{\ip{i}}$;
	\STATE $\bar{\ten{X}}_t = \ten{X}_t
	+ \gamma_t ( \ten{X}_t - \ten{X}_{t - 1} )$;
	\IF{$F_{\tau}(\bar{\ten{X}}_t) \le F_{\tau}(\ten{X}_t)$}
	\STATE $\ten{V}_t = \bar{\ten{X}}_t$, $\gamma_{t + 1} = \min(\frac{\gamma_t}{p}, 1)$;
	\ELSE
	\STATE $\ten{V}_t = \ten{X}_t$, $\gamma_{t + 1} = p \gamma_t$;
	\ENDIF
	\STATE $\ten{Z}_t =\ten{V}_t - \frac{1}{\tau} \SO{\ten{V}_t - \ten{O}}$;
	\\// compute $\SO{\ten{V}_t - \ten{O}}$ using sparse tensor format;
	\FOR{$i = 1, \dots, D$}
	\STATE $\mathbf{X}^i_{t + 1}  = \Px{\frac{\lambda_i}{\tau} \phi} { ( \ten{Z}_t )_{\ip{i}} }$;
	// keep as 
	$\mathbf{U}_{t}^i ( \mathbf{V}_{t}^i )^{\top}$;
	\ENDFOR
	\ENDFOR
	\OUTPUT $\ten{X}_{T + 1}$.
\end{algorithmic}
\label{alg:tcnr}
\end{algorithm}




\subsection{Convergence Analysis} 
\label{sec:convana}

Adaptive momentum has not been used
with the PA algorithm. 
Besides,
previous proofs of the PA algorithm does not involve 
folding/unfolding operations.
Thus,
previous proofs cannot be directly used.
In the following, 
first note that
the proximal step in \eqref{eq:tsplr_3}
implicitly corresponds to a new regularizer.

\begin{proposition}
\label{pr:reg}
There exists a function $\bar{g}$
such that
$\Px{\frac{1}{{\tau}}\bar{g}}{\ten{Z}}
\! = \! \frac{1}{D} \sum_{i = 1}^D
[ \Px{\frac{\lambda_i}{\tau} \phi}
{ \left[ \ten{Z} ]_{\ip{i}} } \right]^{\ip{i}}\!\!$
for any $\tau \! > \! 0$.
\end{proposition}

Let the objective with the new regularizer be
$F_{\tau}(\ten{X}) 
= f(\ten{X}) + \bar{g}(\ten{X})$.
The following bounds the difference between the optimal values 
($F^{\min}$ and 
$F_{\tau}^{\min}$, respectively)
of the objectives $F$ in (\ref{eq:pro}) and $F_{\tau}$.


\begin{proposition}
\label{pr:bnd}
$0 \! \le \! F^{\min} \! - \! F_{\tau}^{\min}
\! \le \! \frac{L^2}{ 2 \tau D} \sum_{d = 1}^D \lambda_d^2$.
\end{proposition}

\subsubsection{With Smooth Assumption}
As in Section~\ref{sec:rel:nonreg},
we assume
that $f$ 
is $L$-Lipschitz smooth.
the following shows that
Algorithm~\ref{alg:tcnr}
converges
to a critical point
(Theorem~\ref{thm:conv})
at the rate of
$O(1/T)$ 
(Corollary~\ref{coro:rate}).
Note that this is the best possible rate for
first-order methods on general nonconvex problems \cite{nesterov2013introductory,ghadimi2016accelerated}.

\begin{theorem} \label{thm:conv}
The sequence $\{ \ten{X}_t \}$ generated from Algorithm~\ref{alg:tcnr} has at least one limit point,
	and all limits points are critical points of $F_{\tau}(\ten{X})$.
\end{theorem}

\begin{corollary}\label{coro:rate}
(i) If $\ten{X}_{t + 1} = \ten{V}_t$,
then $\ten{X}_{t + 1}$ is a critical point of $F_{\tau}$;
(ii) 
let  $\eta = \tau \! - \! \rho \! - \! DL$.
then $\min_{t = 1, \dots, T}\frac{1}{2}\NM{\ten{X}_{t + 1} \! - \! \ten{V}_t}{F}^2 \le
\frac{1}{\eta T }[  F_{\tau}(\ten{X}_1) \! - \! F^{\min}_{\tau} ]$;
\end{corollary}

\begin{remark} \label{rmk:tau}
A larger $\tau$ leads to a better approximation to the original problem $F$
(Proposition~\ref{pr:bnd}).
However, 
it also leads to smaller steps (step~12 in Algorithm~\ref{alg:tcnr})
and thus slower convergence (Corollary~\ref{coro:rate}).
\end{remark}

\subsubsection{With Kurdyka-Lojasiewicz  Condition}
The Kurdyka-Lojasiewicz (KL) condition
\cite{attouch2013convergence,bolte2014proximal} 
has been popularly used in nonconvex optimization, particularly
in gradient \cite{attouch2013convergence} and proximal gradient descent algorithms \cite{bolte2014proximal,li2015accelerated,Li2017ada}.
	
\begin{definition}
\label{def:kl}
A function $h$: $\mathbb{R}^n \rightarrow (-\infty, \infty]$ has
the {\em uniformized
KL property\/}
if for every compact set $\mathcal{S} \in \text{dom}(h)$
on which $h$ is a constant,
there exist $\epsilon$, $c > 0$ such that for 
all $\mathbf{u} \in \mathcal{S}$
and all $\bar{\mathbf{u}} \in \{ \mathbf{u} : \min_{\mathbf{v} \in \mathcal{S} } \NM{\mathbf{u} - \mathbf{v}}{2} \le \epsilon \} \cap
\{ \mathbf{u} 
: f(\bar{\mathbf{u}}) < f(\mathbf{u}) < f(\bar{\mathbf{u}}) + c \}$,
one has 
$\psi'
\left( f(\mathbf{u}) - f(\mathbf{\bar{u}}) \right)
\min_{\mathbf{v} \in \partial f(\mathbf{u})}
\NM{ \mathbf{v} }{F} > 1$,
where 
$\psi(\alpha) = \frac{C}{\beta} \alpha^{\beta}$ for some $C > 0$,
$\alpha \in [0, c)$ and $\beta \in (0, 1]$.
\end{definition}
%

The following extends this to Algorithm~\ref{alg:tcnr}.

\begin{theorem}\label{thm:klrate}
Let $r_t = F_{\tau}(\ten{X}_t) - F_{\tau}^{\min}$.
If $F_{\tau}$ has the uniformized KL property,
for a sufficiently large $t_0$, we have
\begin{itemize}[noitemsep,topsep=0pt,parsep=0pt,partopsep=0pt,leftmargin=13pt]
\item[1.] If $\beta = 1$,
$r_t$ reduces to zero in finite steps;
\item[2.] If $\beta \!\in\! [\frac{1}{2}, 1)$, 
$r_t 
\! \le \! 
( \frac{d_1 C^2}{1 + d_1 C^2} )^{t - t_0} r_{t_0}$ where
$d_1 = \frac{2 (\tau + \rho)^2}{\eta}$;
		
\item[3.] If
$\beta \! \in \! (0, \frac{1}{2})$, 
$r_t \! \le \! ( \frac{C}{(t - t_0)d_2(1 - 2\beta)} )^{1/(1 - 2 \beta)}$
where 
$d_2 = \min
\{ \frac{1}{2 d_1 C}, \frac{C}{1 - 2\beta} (2^{\frac{2\beta - 1}{2\beta - 2}} - 1)
r_{t_0} \}$.
\end{itemize}
\end{theorem}

Though the 
convergence rates in 
Corollary~\ref{coro:rate} and Theorem~\ref{thm:klrate}
are the same 
as when momentum is not used,
the proposed  algorithm does 
have 
faster
convergence empirically,
as will be shown in Section~\ref{sec:exps}.
This also agrees with previous studies showing that adaptive momentum 
can significantly accelerate empirical convergence
\cite{duchi2011adaptive,kingma2014adam,li2015accelerated,Li2017ada,yao2017efficient}.

\section{Experiments} \vspace{-5px}
\label{sec:exps}

\begin{table*}[ht]
	\centering
	\vspace{-10px}
	\caption{Testing RMSE, CPU time and space required for the synthetic data.
		The left is for small $I_3$,
		and the right is for large $I_3$.
		More results with $\bar{c} = 50$ (for small $I_3$) and $\hat{c} = 20$ (for large $I_3$) are in Appendix~\ref{app:imldet}.}
	\label{tab:synperf}
	\scalebox{0.80}{
		\begin{tabular}{c| c | c | c | c || c | c | c}
			\hline
			\multicolumn{2}{c|}{}   &   \multicolumn{3}{c||}{small $I_3$: $\bar{c}=100$, sparsity: $3.09\%$}    &      \multicolumn{3}{c}{large $I_3$: $\hat{c}=40$, sparsity:$2.70\%$}       \\
			\multicolumn{2}{c|}{}   & RMSE                       & space (MB)            & time (sec)           & RMSE                       & space (MB)            & time (sec)             \\ \hline
			convex      & PA-APG & 0.0149$\pm$0.0011          & 302.4$\pm$0.1         & 2131.7$\pm$419.9     & 0.0098$\pm$0.0001          & 4804.5$\pm$598.2      & 6196.4$\pm$2033.4      \\ \hline
			(nonconvex)   & GDPAN  & \textbf{0.0103$\pm$0.0001} & 171.5$\pm$2.2         & 665.4$\pm$99.8       & \textbf{0.0006$\pm$0.0001} & 3243.3$\pm$489.6      & 3670.4$\pm$225.8       \\ \cline{2-8}
			capped-$\ell_1$ & sNORT  & \textbf{0.0103$\pm$0.0001} & \textbf{14.0$\pm$0.8} & 27.9$\pm$5.1         & \textbf{0.0006$\pm$0.0001} & \textbf{44.6$\pm$0.3} & 575.9$\pm$70.9         \\ \cline{2-8}
			& NORT   & \textbf{0.0103$\pm$0.0001} & 14.9$\pm$0.9          & \textbf{5.9$\pm$1.6} & \textbf{0.0006$\pm$0.0001} & 66.3$\pm$0.6          & 89.4$\pm$13.4          \\ \hline
			(nonconvex)   & GDPAN  & 0.0104$\pm$0.0001          & 172.2$\pm$1.5         & 654.1$\pm$214.7      & \textbf{0.0006$\pm$0.0001} & 3009.3$\pm$376.2      & 3794.0$\pm$419.5       \\ \cline{2-8}
			LSP       & sNORT  & 0.0104$\pm$0.0001          & 14.4$\pm$0.1          & 27.9$\pm$5.7         & \textbf{0.0006$\pm$0.0001} & \textbf{44.6$\pm$0.2} & 544.2$\pm$75.5         \\ \cline{2-8}
			& NORT   & 0.0104$\pm$0.0001          & 15.1$\pm$0.1          & \textbf{5.8$\pm$2.8} & \textbf{0.0006$\pm$0.0001} & 62.1$\pm$0.5          & \textbf{81.3$\pm$24.9} \\ \hline
			(nonconvex)   & GDPAN  & 0.0104$\pm$0.0001          & 172.1$\pm$1.6         & 615.0$\pm$140.9      & \textbf{0.0006$\pm$0.0001} & 3009.2$\pm$412.2      & 3922.9$\pm$280.1       \\ \cline{2-8}
			TNN       & sNORT  & 0.0104$\pm$0.0001          & 14.4$\pm$0.1          & 26.2$\pm$4.0         & \textbf{0.0006$\pm$0.0001} & \textbf{44.7$\pm$0.2} & 554.7$\pm$44.1         \\ \cline{2-8}
			& NORT   & \textbf{0.0103$\pm$0.0001} & 15.1$\pm$0.1          & \textbf{5.3$\pm$1.5} & \textbf{0.0006$\pm$0.0001} & 63.1$\pm$0.6          & \textbf{78.0$\pm$9.4}  \\ \hline
	\end{tabular}}
	\vspace{-10px}
\end{table*}

In this section, experiments are performed on both synthetic (Section~\ref{sec:syn}) and real-world data sets (Section~\ref{sec:expreal}).


\subsection{Synthetic Data} \vspace{-5px}
\label{sec:syn}


The setup is as in \cite{song2017tensor}. We first generate
$\bar{\ten{O}} \! = \! \sum_{i = 1}^5 \! s_i 
(  \mathbf{a}_i \! \circ \! \mathbf{b}_i^{} \! \circ \! \mathbf{c}_i )$,
where $\mathbf{a}_i \! \in \! \R^{I_1}$, 
$\mathbf{b}_i \! \in \! \R^{I_2}$ and $\mathbf{c}_i \! \in \! \R^{I_3}$, and
$\circ$ denotes the outer product (i.e., 
$[\mathbf{a} \circ \mathbf{b} \circ \mathbf{c}]_{ijk} = a_i b_j c_k$).
All elements in $\mathbf{a}_i$'s, $\mathbf{b}_i$'s, $\mathbf{c}_i$'s and $s_i$'s are sampled independently from the standard normal
distribution.
This is then corrupted 
by Gaussian noise from $\mathcal{N}(0, 0.01)$
to form $\ten{O}$.
A total of $\NM{\Omega}{1} = I_{+} I_3 \log(I_{\times})/5$ random elements are observed from $\ten{O}$.
We use $50\%$ of them for training, 
and the remaining $50\%$ for validation.
Testing is evaluated on the unobserved elements in $\ten{\bar{O}}$.

Three nonconvex penalties as used:
capped-$\ell_1$ \cite{zhang2010nearly},
LSP \cite{candes2008enhancing} and TNN \cite{hu2013fast}.
The proposed NORT algorithm
is compared with (i)
its slower variant without adaptive momentum 
(denoted sNORT);
(ii)
GDPAN \cite{zhong2014gradient},
which directly applies PA algorithm to \eqref{eq:compopt} as described in \eqref{eq:tsplr_1}-\eqref{eq:tsplr_3}; 
and (iii)
PA-APG \cite{yu2013better},
which solves the convex overlapped nuclear norm minimization problem.
For NORT,
$\tau$ has to be larger than $\rho + D L$
(Corollary~\ref{coro:rate}).
However, 
a large $\tau$ leads to slow convergence
(Remark~\ref{rmk:tau}).
Hence, 
we set $\tau = 1.01(\rho + D L)$.
Moreover,
we set $\gamma_1 = 0.1$ and $p = 0.5$ as in \cite{Li2017ada}.
Besides, $F_{\tau}$ in step~5 of Algorithm~\ref{alg:tcnr} is hard to evaluate, 
and we use $F$ instead as in \cite{zhong2014gradient}.
All algorithms are implemented in Matlab, with sparse tensor and matrix operations in C.  
Experiments are performed on a PC with Intel-i8 CPU and 32GB memory.

Following \cite{lu2016nonconvex,yao2017efficient,yao2018large},
performance is evaluated by (i) root-mean-square-error on the 
unobserved elements:
$\text{RMSE} \! = \! \NM{P_{\bar{\Omega}}(\ten{X} \! - \! \bar{\ten{O}})}{F} \! / \! \NM{ \bar{\Omega} }{1}^{0.5}$,
where $\ten{X}$ is the low-rank tensor recovered,
and $\bar{\Omega}$ contains the unobserved elements in $\bar{\ten{O}}$;
and (ii) CPU time.
To reduce statistical variation, results are averaged over five repetitions.

\subsubsection{Small $I_3$}

Recall that 
for the 3-order tensor considered here, 
we assume 
that $I_1 \!\ge\! I_2 \!\ge\! I_3$.
In this experiment, 
we first study the case where
$I_3$ 
is small.
We set $I_1 \! = \! I_2 \! = \! 25 \bar{c}$, 
where $\bar{c} \! = \! 100 $,
$I_3 \! = \! 5$,
and $D \! = \! 2$.

\begin{figure}[ht]
	\centering
	\vspace{-10px}
	\subfigure[vs iterations.]
	{\includegraphics[width = 0.49\columnwidth]{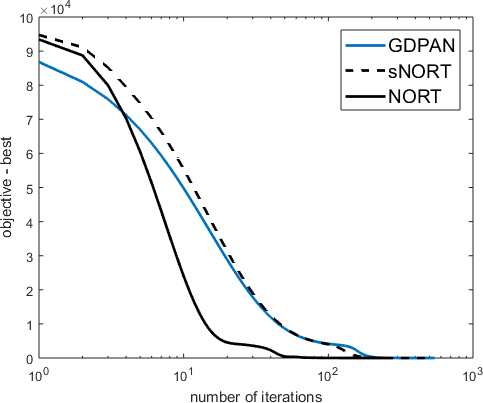}}
	\subfigure[vs time.]
	{\includegraphics[width = 0.485\columnwidth]{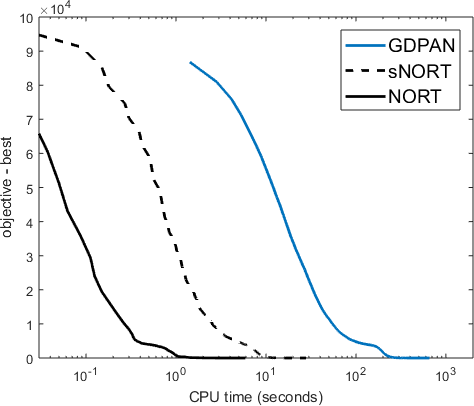}}
	\vspace{-10px}
\caption{Convergence of the objective on the synthetic data for
small 
$I_3$ 
(capped-$\ell_1$ regularizer).}
	\label{fig:syn2}
	\vspace{-5px}
\end{figure}

Table~\ref{tab:synperf} shows
the results\footnote{In all tables, the best and comparable performances (according to the pairwise t-test with 95\% confidence) 
	are highlighted.}.
As can be seen, PA-APG,
which is based on the convex overlapped nuclear norm,
has much higher testing RMSEs than
those with nonconvex regularization.
Besides,
the various nonconvex penalties
(capped-$\ell_1$, LSP and TNN)
have similar empirical testing RMSEs, 
as is also observed in \cite{lu2016nonconvex,yao2017efficient,yao2018large}.
As for space,
NORT and its variant sNORT
need much less memory than PA-APG and GDPAN, 
as they do not explicitly construct
dense tensors 
during the iterations.
As for time, 
NORT is the fastest. 
Figure~\ref{fig:syn2} shows
convergence of the objective value.\footnote{Plots for LSP and TNN are similar, and
so are not shown because of the lack of space.}
As can be seen, sNORT and GDPAN have similar speeds w.r.t. the number of iterations.
However,
sNORT is much faster 
when measured against time,
as it utilizes the ``sparse plus low-rank'' structure.
NORT is even faster due to usage of adaptive momentum.

\subsubsection{Large $I_3$} 

\begin{figure}[ht]
	\centering
	\subfigure[vs iterations.]
	{\includegraphics[width = 0.485\columnwidth]{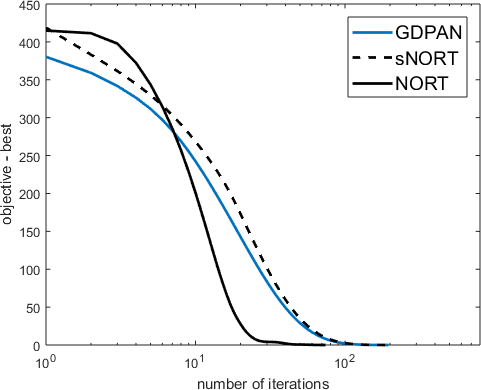}}
	\subfigure[vs time.]
	{\includegraphics[width = 0.49\columnwidth]{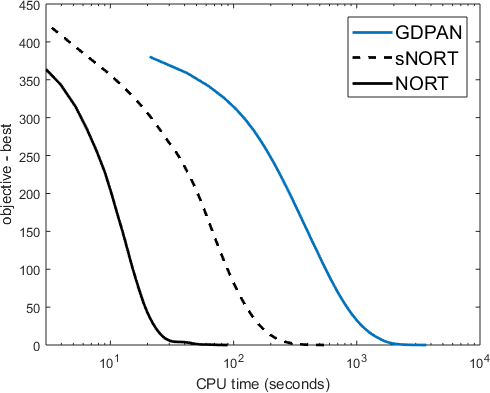}}
	
	\vspace{-10px}
	
	\caption{Convergence of the objective on synthetic data for 
	large
	$I_3$ 
		(capped-$\ell_1$ regularizer).}
	\label{fig:syn3}
	\vspace{-10px}
\end{figure}

In this experiment, we set
 $I_1 = I_2 = I_3 = 10 \hat{c}$,
where $\hat{c} =  40$;
and $D \! = \! 3$ is used here.
Results are shown in Table~\ref{tab:synperf}.
Again, the
capped-$\ell_1$, LSP and TNN regularizers
yield the same RMSE.
GDPAN, sNORT and NORT
all have much lower RMSEs than PA-APG.
Convergence of the objective value is shown in Figure~\ref{fig:syn3}.
Again,
NORT is the fastest,
and GDPAN 
is the slowest.
NORT and sNORT need much less memory than GDPAN.

\textit{3). $D\!=\!2$ vs $D\!=\!3$.}
Recall that $D$ in \eqref{eq:pro} can be either $2$ and $3$.
We expect 
$D \! = \! 2$ to be better when $I_3$ is small, and vice versa. 
This will be verified in this section.
The setup is as in Sections~4.1.1 and 4.1.2.
We only experiment with NORT, 
as the other baselines are less efficient.

Results are shown in Table~\ref{tab:compdiffD}.
As can be seen, 
when $I_3$ is small, 
$D \! = \! 2$ and $3$ yield similar RMSEs.
However,
$D$ = $3$ is much slower than $D \! = \! 2$.
When $I_3$ is small,
the third mode
is not 
low-rank.
Thus,
the proximal step for the third mode 
is much more expensive than those for the first two modes as
we cannot have $k_t^3 \ll I_3$.
Moreover,
$D$ = $3$ requires much larger
space than $D \! = \! 2$.

\begin{table}[ht]
\centering
\vspace{-10px}
\caption{NORT with different $D$'s in \eqref{eq:pro} on the synthetic data.}
\label{tab:compdiffD}
\scalebox{0.80}{
\begin{tabular}{c| c | c | C{25px} | C{25px} | c | C{25px} | C{25px} }
	\hline
	          &     & \multicolumn{3}{c|}{small $I_3$ ($\bar{c}=100$)} & \multicolumn{3}{c}{large $I_3$ ($\hat{c}=40$)}  \\
	          & $D$ & RMSE            & space (MB)    & time (sec)     & RMSE            & space (MB)    & time (sec)    \\ \hline
	 capped   & 2   & \textbf{0.0103} & \textbf{14.0} & \textbf{5.9}   & 0.0009          & 46.7         & \textbf{40.0} \\ \cline{2-8}
	-$\ell_1$ & 3   & \textbf{0.0103} & 78.7          & 918.7          & \textbf{0.0006} & \textbf{66.3} & 89.4          \\ \hline
	   LSP    & 2   & \textbf{0.0104} & \textbf{14.1} & \textbf{5.8}   & 0.0010          & 45.2         & 50.8          \\ \cline{2-8}
	          & 3   & \textbf{0.0103} & 78.7          & 899.7          & \textbf{0.0006} & \textbf{62.1} & 81.3          \\ \hline
	   TNN    & 2   & \textbf{0.0103} & \textbf{14.4} & \textbf{5.3}   & 0.0009          & 46.8         & \textbf{39.3} \\ \cline{2-8}
	          & 3   & \textbf{0.0104} & 77.8          & 615.5          & \textbf{0.0006} & \textbf{63.1} & 78.0          \\ \hline
\end{tabular}}
\end{table}


When $I_3$ is large, 
it is slightly more expensive on CPU time and space.
However,
$D \! = \! 2$ has much worse 
testing RMSE
than $D \! = \! 3$,
as it cannot capture the low-rank property on the third mode.



\subsection{Real-World Data Sets}
\label{sec:expreal}

In this section, we perform evaluation on 
color images,
remote sensing data,
and 
social network data.
As different nonconvex regularizers have similar performance, we will only use LSP. 
Moreover, 
based on the observations in Section~4.1.3,
we use $D \! = \! 2$ if $I_3$ is $\le \! 10$,
and $D \! = \! 3$
otherwise.
Besides comparing with GDPAN,
the proposed 
NORT algorithm
is also 
compared with:\footnote{sNORT is not compared as it has been shown to be slower than NORT.  Neither do we compare with \cite{bahadori2014fast}, which is inferior to FFW above \cite{guo2017efficient}; and \cite{rauhut2017low}, whose its code is not publicly available and the more recent TMac-TT solves the same problem.}
(i) 
Algorithms for various convex regularizers
including
ADMM
\cite{tomioka2010estimation},
FaLRTC \cite{liu2013tensor},
PA-APG \cite{yu2013better},
FFW \cite{guo2017efficient},
TenNN \cite{zhang2017exact},
and TR-MM \cite{nimishakavi2018dual};
(ii) Factorization-based algorithms
including
RP \cite{kasai2016low},
TMac \cite{xu2013tmac},
CP-WOPT \cite{acar2011scalable}, and
TMac-TT \cite{bengua2017efficient}.
More details about these methods can be found in Table~\ref{tab:comparedalgs} of the Appendix.



\subsubsection{Color Images}
\label{sec:image}

%
%
%

We use the images
\textit{windows},
\textit{tree}
and
\textit{rice} from \cite{hu2013fast},
which are resized to $1000 \times 1000 \times 3$.
Sample images are shown in Appendix~\ref{app:cimg}.
Each pixel is normalized to $[0, 1]$.
We randomly sample 10\% of the pixels for training, which are then  corrupted by
Gaussian noise $\mathcal{N}(0, 0.01)$.
Half of the training pixels are used for validation.
The remaining unseen clean pixels are used for testing.
Performance is measured by
the testing RMSE and CPU time.
To reduce statistical variation, results are averaged over five repetitions.

\begin{figure}[ht]
	\centering
	\vspace{-10px}
	\subfigure[\textit{rice}.]
	{\includegraphics[width = 0.49\columnwidth]{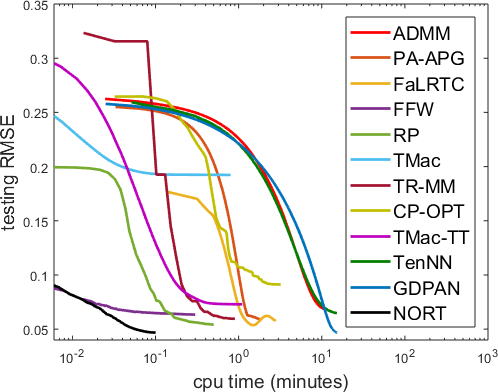}}
	\subfigure[\textit{tree}.]
	{\includegraphics[width = 0.485\columnwidth]{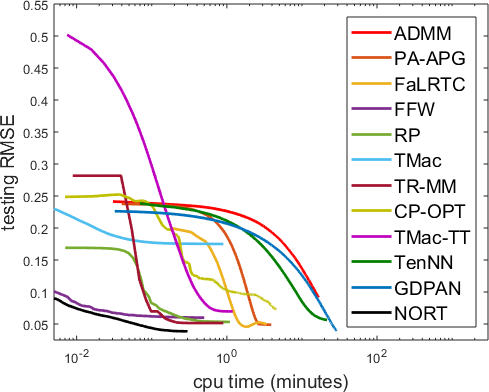}}
	\vspace{-10px}
	\caption{Testing RMSE vs CPU time (seconds) on color images.
		The plot for \textit{windows} is similar, and thus not shown.}
	\label{fig:colorimg}
	\vspace{-10px}
\end{figure}

Table~\ref{tab:colorimg} shows the
results. As can be seen,
the best convex methods 
(PA-APG and FaLRTC)
are
based on the overlapping nuclear norm.
This agrees with our motivation to build a nonconvex regularizer
based on the overlapping nuclear norm.
GDPAN and NORT have similar RMSEs,
that are lower than those by convex regularization
and factorization approach.
Convergence of the testing RMSE is shown in Figure~\ref{fig:colorimg}.
As can be seen,
while ADMM solves the same convex model as PA-APG and FaLRTC,
it 
has 
slower convergence.
FFW, RP and TR-MM are very fast but their testing RMSEs are higher than that of NORT.
By utilizing the ``sparse plus low-rank'' structure and adaptive momentum,
NORT is more efficient than GDPAN.

\begin{table}[ht]
	\centering
	\vspace{-15px}
	\caption{Testing RMSEs ($\times 10^{-1}$) 
	on color images.}
	\scalebox{0.80}{
		\begin{tabular}{c | c| c | c | c}
			\hline
			\multicolumn{1}{c|}{} &         & \textit{rice}            & \textit{tree}            & \textit{windows}         \\ \hline
			       convex         & ADMM    & 0.680$\pm$0.003          & 0.915$\pm$0.005          & 0.709$\pm$0.004          \\ \cline{2-5}
			                      & PA-APG  & 0.583$\pm$0.016          & 0.488$\pm$0.007          & 0.585$\pm$0.002          \\ \cline{2-5}
			                      & FaLRTC  & 0.576$\pm$0.004          & 0.494$\pm$0.011          & 0.567$\pm$0.005          \\ \cline{2-5}
			                      & FFW     & 0.634$\pm$0.003          & 0.599$\pm$0.005          & 0.772$\pm$0.004          \\ \cline{2-5}
			                      & TR-MM   & 0.596$\pm$0.005          & 0.515$\pm$0.011          & 0.634$\pm$0.002          \\ \cline{2-5}
			                      & TenNN   & 0.647$\pm$0.004          & 0.562$\pm$0.004          & 0.586$\pm$0.003          \\ \hline
			       factor-        & RP      & 0.541$\pm$0.011          & 0.524$\pm$0.010          & 0.388$\pm$0.026          \\ \cline{2-5}
			       ization        & TMac    & 1.923$\pm$0.005          & 1.750$\pm$0.006          & 1.313$\pm$0.005          \\ \cline{2-5}
			                      & CP-OPT  & 0.912$\pm$0.086          & 0.733$\pm$0.060          & 0.964$\pm$0.102          \\ \cline{2-5}
			                      & TMac-TT & 0.729$\pm$0.022          & 0.697$\pm$0.147          & 1.045$\pm$0.107          \\ \hline
			        noncvx          & GDPAN   & \textbf{0.467$\pm$0.002} & \textbf{0.388$\pm$0.012} & \textbf{0.296$\pm$0.007} \\ \cline{2-5}
			                      & NORT  & \textbf{0.468$\pm$0.001} & \textbf{0.386$\pm$0.009} & \textbf{0.297$\pm$0.007} \\ \hline
		\end{tabular}
	}
	\label{tab:colorimg}
\end{table}

\subsubsection{Remote Sensing Data}

\begin{figure}[ht]
	\centering
	\subfigure[\textit{Cabbage}.
	\label{fig:cabbage}]
	{\includegraphics[width = 0.49\columnwidth]{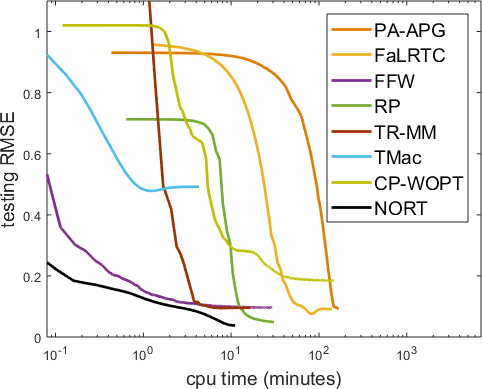}}
	\subfigure[\textit{Female}.
	\label{fig:leresfemale}]
	{\includegraphics[width = 0.49\columnwidth]{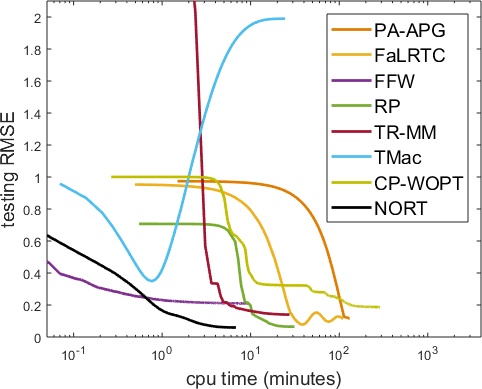}}
	
	\vspace{-10px}
	
	\caption{Testing RMSE vs CPU time (minutes) on remote sensing data.
		The plot for \textit{Scene41} is similar, thus is not shown.}
	\label{fig:hyper}
	\vspace{-5px}
\end{figure}

Experiments are performed on three hyper-spectral data:
\textit{Cabbage} (1312$\times$432$\times$49),
\textit{Scene} (1312$\times$951$\times$49)
and
\textit{Female} (592$\times$409$\times$148).
Details are in Appendix~\ref{app:himg}.
The third dimension is for the bands of images.
We use the same setup as in Section~\ref{sec:image}.
ADMM, TenNN, GDPAN,
and TMac-TT 
are 
slow
and so not compared.
Results are shown in Table~\ref{tab:hypspec}.
Again,
NORT achieves much lower testing RMSE than convex regularization and factorization approach.
Figure~\ref{fig:hyper} shows convergence of
the testing RMSE.
As can be seen,
NORT is fast.

\begin{table}[ht]
	\centering
	\vspace{-10px}
	\caption{Testing RMSEs ($\times {10}^{-2}$) on remote sensing data.}
	\scalebox{0.80}{
		\begin{tabular}{c | c | C{54px} | C{54px} | C{54px} }
			\hline
			        &        & \textit{Cabbage}       & \textit{Scene}          & \textit{Female}        \\ \hline
			convex  & PA-APG & 9.13$\pm$0.06          & 19.65$\pm$0.02          & 11.57$\pm$0.03         \\ \cline{2-5}
			        & FaLRTC & 9.09$\pm$0.02          & 19.20$\pm$0.01          & 11.33$\pm$0.04         \\ \cline{2-5}
			        & FFW    & 9.62$\pm$0.04          & 20.37$\pm$0.02          & 20.96$\pm$0.06         \\ \cline{2-5}
			        & TR-MM  & 9.59$\pm$0.01          & 19.65$\pm$0.02          & 13.97$\pm$0.06         \\ \hline
			factor- & RP     & 4.91$\pm$0.11          & 18.04$\pm$0.05          & 6.47$\pm$0.03          \\ \cline{2-5}
			ization & TMac   & 49.19$\pm$0.59         & 59.70$\pm$0.29          & 198.97$\pm$0.06        \\ \cline{2-5}
			        & CP-OPT & 18.46$\pm$5.14         & 48.11$\pm$0.82          & 18.68$\pm$0.13         \\ \hline
			 noncvx   & NORT & \textbf{3.76$\pm$0.04} & \textbf{17.14$\pm$0.12} & \textbf{5.92$\pm$0.02} \\ \hline
		\end{tabular}}
	\label{tab:hypspec}
\end{table}

\subsubsection{Social Networks}

\begin{figure}[ht]
	\centering
	\subfigure[Subset.]{\includegraphics[width = 0.48\columnwidth]{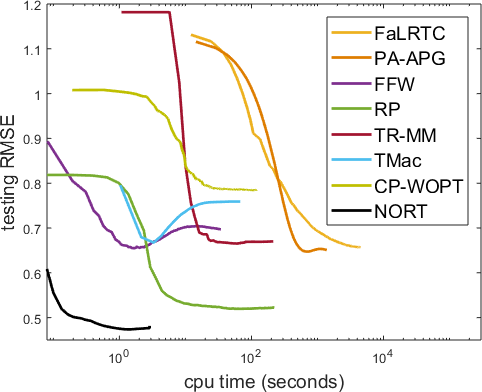}}
	\subfigure[Full set.]{\includegraphics[width = 0.49\columnwidth]{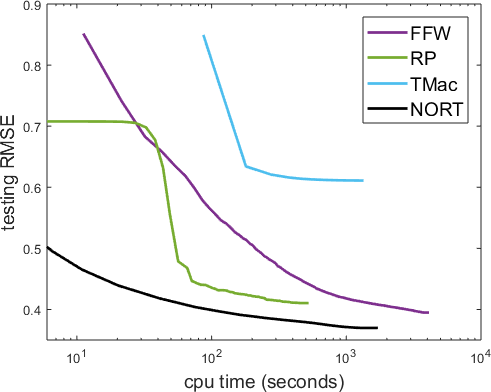}}
	\vspace{-10px}
	\caption{Testing RMSE vs CPU time (seconds) on \textit{Youtube}.}
	\label{fig:linkpred}
	\vspace{-10px}
\end{figure}

In this section,
we
perform 
multi-relational link prediction 
\cite{guo2017efficient}
as a tensor completion problem
on the \textit{YouTube} data set
\cite{lei2009analysis}.
It contains 15,088 users, and describes five types of user interactions.
Thus, it forms a
15088$\times$15088$\times$5 tensor,
with a total of 27,257,790 nonzero elements.
Besides the full set,
we also experiment with a \textit{YouTube} subset
obtained by randomly selecting 1,000 users (leading to 12,101 observations).
We use $50\%$ of the observations for training, another $25\%$ for validation and the rest for testing.
Experiments are repeated five times.
Table~\ref{tab:linkpred} shows the testing RMSE, 
and Figure~\ref{fig:linkpred} shows the convergence.
As can be seen, 
NORT achieves low RMSE and is also much faster.

\begin{table}[ht]
\centering
\vspace{-10px}
\caption{Testing RMSEs on \textit{Youtube} data set.  FaLRTC, PA-APG, TR-MM and CP-OPT are slow, and thus not run on the full set.}
	\scalebox{0.80}
	{	\begin{tabular}{c | C{50px} | C{62px} | C{62px}  }
		\hline
		                        &        & subset                   & full set                 \\ \hline
		\multirow{3}{*}{convex} & FaLRTC & 0.657$\pm$0.060          & ---                      \\ \cline{2-4}
		                        & PA-APG & 0.651$\pm$0.047          & ---                      \\ \cline{2-4}
		                        & FFW    & 0.697$\pm$0.054          & 0.395$\pm$0.001          \\ \cline{2-4}
		                        & TR-MM  & 0.670$\pm$0.098                         & ---                      \\ \hline
		                        & RP     & 0.522$\pm$0.038          & 0.410$\pm$0.001          \\ \cline{2-4}
		     factorization      & TMac   & 0.795$\pm$0.033          & 0.611$\pm$0.007          \\ \cline{2-4}
		                        & CP-OPT & 0.785$\pm$0.040          & ---                      \\ \hline
		         nonconvex           & NORT & \textbf{0.482$\pm$0.030} & \textbf{0.370$\pm$0.001} \\ \hline
	\end{tabular}}
	\label{tab:linkpred}
\end{table}

\section{Conclusion}
\label{sec:concl}

In this paper,
we propose a low-rank tensor completion model with nonconvex regularization.
An efficient
nonconvex proximal average algorithm
is developed,
which maintains the ``sparse plus
low-rank" structure throughout the iterations and  also
incorporates adaptive momentum.
Convergence to critical points is guaranteed.
Experimental results show that the proposed 
algorithm is more efficient
and more accurate
than existing approaches.


{
\small
\bibliography{bib}
\bibliographystyle{icml2019}
}

\end{document}